%% file: main.tex
\documentclass{article} 
\usepackage{iclr2027_conference,times}

\iclrfinalcopy

\input{math_commands.tex}

\usepackage{hyperref}
\usepackage{url}

\usepackage[most]{tcolorbox}
\usepackage{listings}               
\tcbuselibrary{listings,breakable,skins}
\newtcblisting{codeblock}{
  listing only,
  breakable,
  colback=gray!8,
  colframe=gray!55,
  boxrule=0.5pt,
  arc=3pt,
  left=6pt,right=6pt,top=6pt,bottom=6pt,
  listing options={
    basicstyle=\ttfamily\small,
    columns=fullflexible,
    keepspaces=true,
    showstringspaces=false,
    breaklines=true,
    tabsize=2
  }
}

\usepackage{enumitem}
\usepackage{color}
\usepackage{xcolor}  
\usepackage{colortbl}

\usepackage{microtype}
\usepackage{graphicx}
\usepackage{subcaption}
\usepackage{booktabs} 
\usepackage{amssymb}
\usepackage{algorithm}
\usepackage{algorithmic}
\usepackage{amsmath}
\usepackage{amsthm}
\usepackage{multirow}

\newtheorem{theorem}{Theorem}[section]

\usepackage{wrapfig}

\title{Towards Generalizable Reasoning: Group Causal Counterfactual Policy Optimization for LLMs}

\author{Jingyao Wang\textsuperscript{1,2},\, Xingyu Zhang\textsuperscript{1,2},\, Peizheng Guo\textsuperscript{1,2},\, Jiahuan Zhou\textsuperscript{3},\, Zhenghao Zhang\textsuperscript{4},\\ \textbf{Changwen Zheng\textsuperscript{1,2},\, Chuxiong Sun\textsuperscript{1,2},\, Wenwen Qiang\textsuperscript{1,2}}\,\thanks{Corresponding author}\\
\textsuperscript{1}National Key Laboratory of Space Integrated Information System,\\ \,Institute of Software Chinese Academy of Sciences,\\
  \textsuperscript{2}University of Chinese Academy of Sciences,\\
  \textsuperscript{3}Wangxuan Institute of Computer Technology, Peking University,\\
  \textsuperscript{4}Alibaba Group\\
}

\begin{document}

\maketitle

\begin{abstract}
Large language models (LLMs) excel at complex tasks with advances in reasoning capabilities. However, existing reward mechanisms remain tightly coupled to final correctness and pay little attention to the underlying reasoning process: trajectories with sound reasoning but wrong answers receive low credit, while lucky guesses with flawed logic may be highly rewarded, affecting reasoning generalization. 
  From a causal perspective, we interpret multi-candidate reasoning for a fixed question as a family of counterfactual experiments. Building on this, we propose \textbf{G}roup \textbf{C}ausal \textbf{C}ounterfactual \textbf{P}olicy \textbf{O}ptimization (\textbf{G$\mathbf{C^2}$PO}) to explicitly train LLMs to learn generalizable reasoning patterns. It proposes an episodic causal counterfactual reward that jointly captures (i) robustness, encouraging the answer distribution induced by a reasoning step to remain stable under counterfactual perturbations; and (ii) effectiveness, enforcing sufficient variability so that the learned reasoning strategy can transfer across questions. We then construct token-level advantages from this reward and optimize the policy, encouraging LLMs to favor reasoning patterns that are process-valid and counterfactually robust. 
  Extensive experiments on diverse benchmarks demonstrate its advantages.
\end{abstract}


\section{Introduction}

Large Language Models (LLMs) have evolved from handling basic natural language processing tasks to tackling complex reasoning problems \citep{jimenez2024swebenchlanguagemodelsresolve,zhou2025sweetrltrainingmultiturnllm,sadik2025benchmarkingllmcodesmells}. While pre-training establishes a foundation, realizing their full potential, particularly in specialized reasoning domains, requires targeted post-training adjustments \citep{tie2025survey}. Reinforcement Learning (RL) has emerged as a dominant paradigm to enhance reasoning capabilities \citep{shao2024deepseekmath,zhang2024rest,wang2025insightsverificationtrainingverilog}. Among RL-based post-training approaches, Group Relative Policy Optimization (GRPO) \citep{shao2024deepseekmath} and its variants \citep{zhang2025gvpo,gu2025group,wang2025learning} have garnered significant attention by reducing computational overhead while delivering substantial performance gains.

\begin{figure*}[t]
    \centering
    \includegraphics[width=\linewidth]{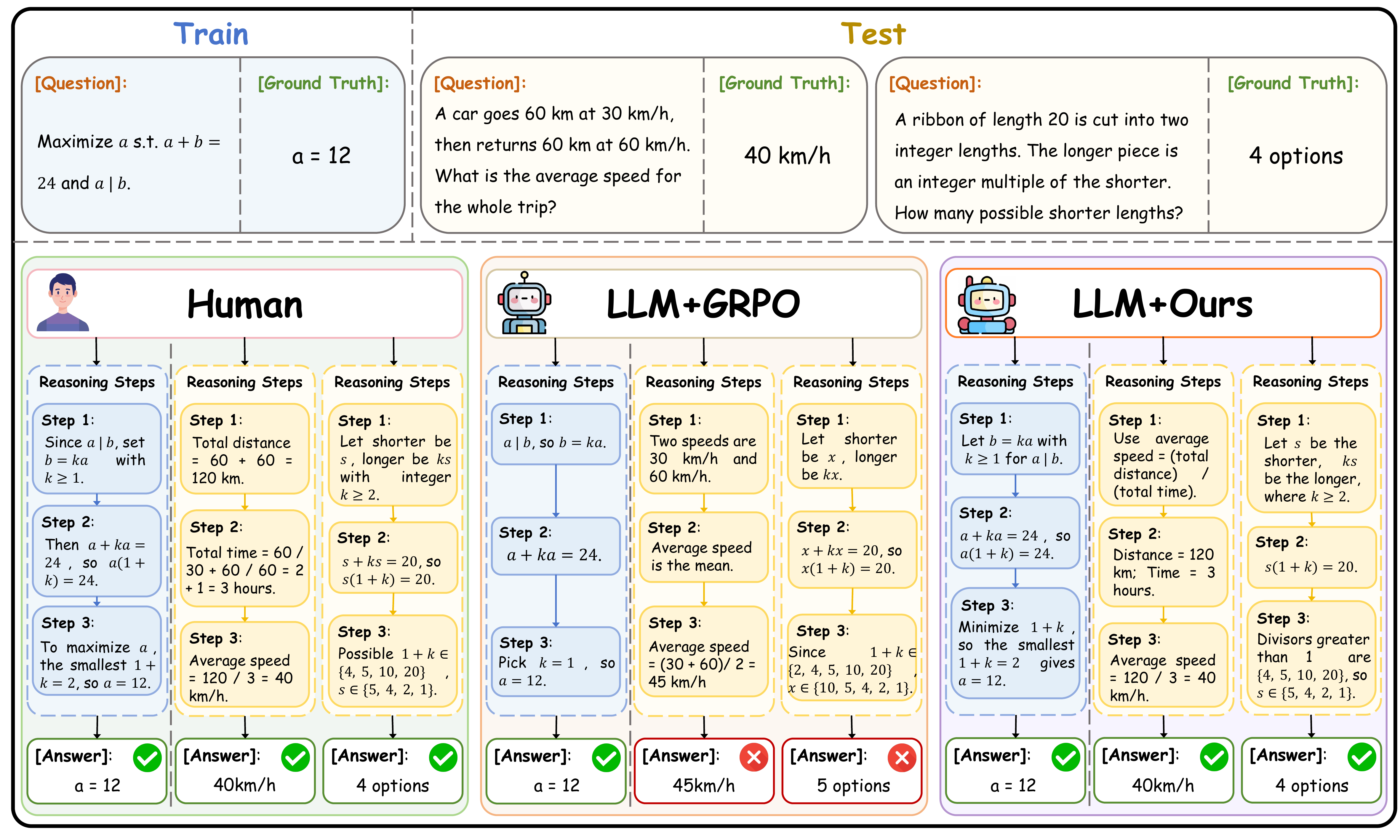}
    \vspace{-0.15in}
    \caption{Example of reasoning trajectories. It illustrates the differences between the reasoning of Human, LLM+GRPO, and LLM+CCPO.}
    \label{fig:intro}
    \vspace{-0.15in}
\end{figure*}

Despite these advancements, our empirical analysis reveals that existing methods may struggle to achieve generalizable reasoning: the models frequently overfit to in-distribution benchmarks and fail to answer unseen questions correctly, e.g., rephrasings, distractors, or harder variants (\textbf{Figure \ref{fig:intro}}).  
Through a toy experiment (\textbf{Subsection \ref{sec:empirical_evidence}}), we trace this generalization failure to a misalignment in reward supervision, i.e., current algorithms heavily reward the final answer rather than the fidelity of the reasoning process. It inadvertently incentivizes LLMs to mimic answers and exploit superficial shortcuts rather than learning transferable reasoning principles.
Specifically, current approaches primarily rely on outcome-based binary rewards \citep{shao2024deepseekmath,yeo2025demystifyinglongchainofthoughtreasoning,yang2025thinkneedselfadaptivechainofthought}, assigning 1 to correct answers and 0 otherwise, regardless of the logical validity of the intermediate steps. 
Although some works use Process Reward Models (PRMs) \citep{wang2025learning,zhang2025lessons,tan2025gtpo} to score intermediate steps, they are typically trained on chain-of-thought (CoT) traces filtered by final correctness, while abundant incorrect samples are treated as uniformly negative \citep{chen2025pass,ye2025beyond}. 
Thus, they may inherit the outcome bias: ``lucky guesses'' driven by flawed logic receive undue reinforcement, whereas rigorous reasoning trails derailed by minor terminal slips are blindly penalized.
To address this issue, we aim to answer a core question: \emph{How can reward signals distinguish genuine reasoning from lucky guesses?}

From a causal perspective \citep{pearl2009causality}, we propose that this distinction hinges on causal invariance: genuine reasoning captures stable and transferable mechanisms that persist across context shifts, whereas lucky guesses exploit spurious correlations that happen to yield correct answers only by chance.
As illustrated in \textbf{Figure \ref{fig:scm_motivation}}, generalizable reasoning relies on invariant structures (e.g., core constraints, correct variable binding, etc.) that remain valid across diverse problem formulations, rather than spurious cues (e.g., seeing ``average'' and mechanically applying a fixed formula) that only correlate with success in a narrow distribution (\textbf{Subsection \ref{sec:motivation_analysis}}). 
Crucially, within the GRPO paradigm, the multiple trajectories sampled per prompt naturally provide a testbed of counterfactual trials (\textbf{Theorem \ref{theorem:motivation}}). By evaluating process behaviors across these parallel rollouts, we can isolate logically consistent transitions that signify genuine causal invariance.
Motivated by this, we propose two complementary principles for reward design: (i) robustness, which assesses whether an intermediate step maintains stability under local perturbations, verifying the presence of an invariant mechanism; and (ii) effectiveness, which ensures the step retains sufficient task-relevant information, preventing the representations from collapsing into trivial states. Together, they establish a process-level supervisory signal that reliably credits sound reasoning while penalizing lucky guesses, even when final answers mislead, thereby steering LLMs toward robust reasoning generalization.

Building on these insights, we propose Group Causal Counterfactual Policy Optimization (G$C^2$PO). It introduces an episodic causal counterfactual reward to explicitly evaluate intermediate reasoning. This reward couples a stability term measuring causal invariance under perturbation with an expressiveness term that penalizes uninformative representations via optimal decay rates. 
G$C^2$PO operates in three steps: (i) Episode segmentation (\textbf{Subsection \ref{sec:method_episode_segmentation}}): we automatically break down solution paths into several reasoning episodes; (ii) Reward calculation (\textbf{Subsection \ref{sec:method_reward}}): for each episode, we construct local perturbations in the representation space and calculate the reward with a Monte Carlo estimator, jointly estimating robustness and effectiveness; (iii) Policy optimization (\textbf{Subsection \ref{sec:method_policy}}): we combine episodic rewards with outcome rewards to allocate token-level advantages, encouraging the LLMs towards generalizable reasoning.
Extensive experiments demonstrate its advantages.

\textbf{The main contributions are as follows:} 
(i) We reveal a structural bias in existing post-training through empirical and causal analyses, demonstrating how current reward mechanisms entangle process validity with final correctness and hinder generalization.
(ii) We propose G$C^2$PO, a novel post-training framework that utilizes episodic causal counterfactual rewards to evaluate the robustness and effectiveness of underlying reasoning patterns.
(iii) Extensive experiments across various benchmarks and models demonstrate the superiority of G$C^2$PO, improving the generalization of LLM reasoning.


\section{Problem Settings and Analyses}

\subsection{Problem Settings}
\label{sec:problem_settings}

Our goal is to fine-tune a policy $\pi_\theta$ (i.e., LLM) to answer arbitrary questions. 
Taken GRPO \citep{shao2024deepseekmath} as an example: for each question $x\sim \mathcal{D}$, we sample $K$ candidate trajectories $\{y_k\}_{k=1}^K \sim \pi_{\theta_{\mathrm{old}}}(\cdot | x)$ using the old policy $\pi_{\theta_{\rm old}}$. For each $y_k$, we compute the outcome reward $R_{\mathrm{out}}(x,y_k) \in \{0,1\}$ using an automatic verifier (e.g., correctness plus formatting). Next, we calculate:
\begin{equation}
\begin{aligned}
    \scalebox{0.95}{$A_k(x) = \frac{r_k - \bar r(x)}{\sqrt{s_r^2(x)}},$} \text{s.t.}\; \scalebox{0.95}{$\bar r(x) = \frac{1}{K}\sum_{k=1}^K r_k,\;
s_r^2(x) = \frac{1}{K}\sum_{k=1}^K \big(r_k - \bar r(x)\big)^2,$}
\end{aligned}
\end{equation}
where $A_k(x)$ is the group-relative advantage of trajectory $y_k$, $r_k$ is the trajectory-level scalar reward, and $\bar r(x)$ and $s_r^2(x)$ are the empirical mean and variance of the rewards across the $K$ sampled trajectories for question $x$, respectively.
By distributing this advantage uniformly over the tokens of $y_k$, we then optimize $\pi_\theta$ by maximizing:
\begin{equation}\label{eq:problem_grpo}
\begin{gathered}
    \scalebox{0.85}{$\mathcal{J}_{\mathrm{GRPO}}(\theta)=\mathbb{E}_{x \sim \mathcal{D},\, \{y_k\} \sim \pi_{\theta_{\mathrm{old}}}(\cdot \mid x)}\Big[\frac{1}{K}\sum_{k=1}^K \frac{1}{T_k}\sum_{t=1}^{T_k}\min \Big(\rho_{k,t}(\theta)A_k(x),\tilde\rho_{k,t}(\theta)A_k(x)\Big)
    -\beta_{\mathrm{KL}}\,\mu_{\mathrm{KL}}\big(\pi_\theta\,\Vert\,\pi_{\mathrm{ref}}\big)\Big],$} \\
    \text{s.t.}\;\scalebox{0.95}{$\rho_{k,t}(\theta)=\frac{\pi_\theta\big(y_{k,t} \mid x, y_{k,<t}\big)}{\pi_{\theta_{\mathrm{old}}}\big(y_{k,t} \mid x, y_{k,<t}\big)},$}\;
    \scalebox{0.9}{$\tilde\rho_{k,t}(\theta)=\operatorname{clip}\big(\rho_{k,t}(\theta),1-\epsilon,1+\epsilon\big),$} \\
    \scalebox{0.98}{$\mu_{\mathrm{KL}}\big(\pi_\theta\,\Vert\,\pi_{\mathrm{ref}}\big)
    =\frac{\pi_{\mathrm{ref}}(y_{k,t}\mid x,y_{k,<t})}{\pi_{\theta}(y_{k,t}\mid x,y_{k,<t})}
    -\log \frac{\pi_{\mathrm{ref}}(y_{k,t}\mid x,y_{k,<t})}{\pi_{\theta}(y_{k,t}\mid x,y_{k,<t})}-1,$}
\end{gathered}
\end{equation}
where $\epsilon$ is the clipping radius, $\beta_{\mathrm{KL}}$ controls KL regularization $\mu_{\mathrm{KL}}$, and $\pi_{\mathrm{ref}}$ is the reference policy.

\begin{figure*}[t]
    \centering
    \includegraphics[width=\linewidth]{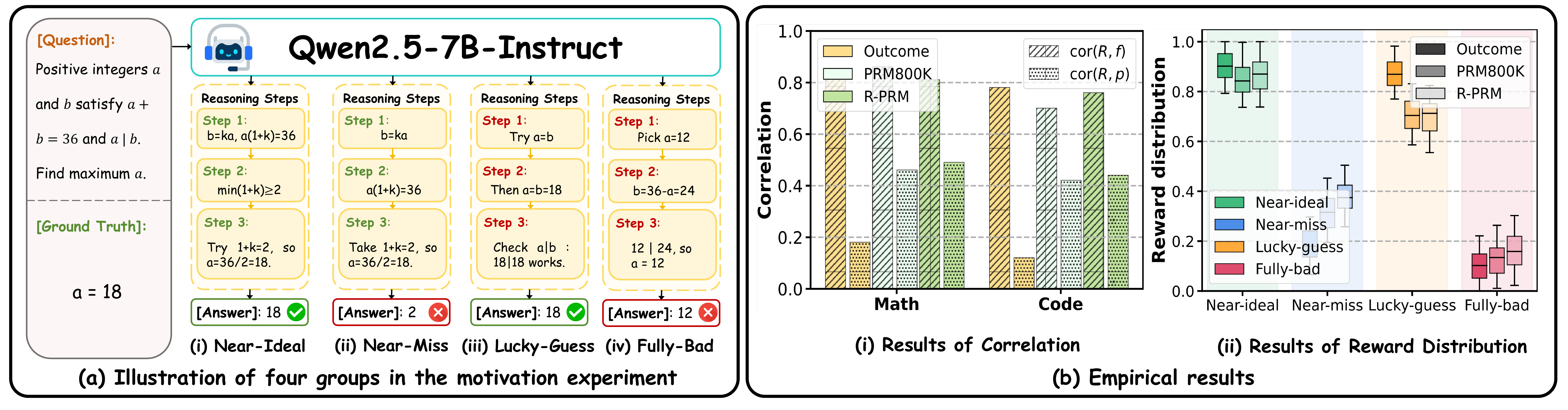}
    \vspace{-0.15in}
    \caption{(a) Examples of four trajectory groups (\textbf{Subsection \ref{sec:empirical_evidence})}. (b) Empirical results of the motivation experiment. See \textbf{Appendix D.4} for details.}
    \label{fig:intro}
    \vspace{-0.15in}
\end{figure*}

\subsection{Empirical Evidence}
\label{sec:empirical_evidence}
Despite strong performance, existing reward mechanisms are tightly coupled to final correctness rather than the reasoning process \citep{shao2024deepseekmath,khalifa2025process,tan2025gtpo}, affecting generalization. To validate this, we conduct an experiment for analysis.
Specifically, we construct an evaluation set from GSM8K \citep{cobbe2021training}, MATH \citep{hendrycks2021measuring}, and HumanEval \citep{chen2021evaluating}, and use Qwen2.5-7B-Instruct to generate multiple trajectories per question. For each trajectory $y_k$, we record: (i) final correctness $f_k \in \{0,1\}$, determined by symbolic solver or code execution; and (ii) process validity $p_k \in [0,1]$, obtained from independent annotations of intermediate steps (human raters assisted by arithmetic checkers and code execution). 
We split the trajectories into four groups (\textbf{Figure~\ref{fig:intro}(b)}) by evaluating whether their reasoning process and answer are correct, i.e., near ideal, near-miss, lucky guess, and fully bad.
We then apply two families of reward mechanisms: (i) the outcome reward in GRPO \citep{shao2024deepseekmath}; and (ii) representative PRMs, i.e., Qwen2.5-Math-7B-PRM800K and R-PRM-7B. We evaluate: (i) the correlation $\operatorname{cor}(R(\cdot), f)$ and $\operatorname{cor}(R(\cdot), p)$; and (ii) the rewards of different models over four groups.

As shown in \textbf{Figure~\ref{fig:intro}(c)}, (i) existing methods align much more strongly with final correctness than with process validity: $\operatorname{cor}(R(\cdot), f)$ is much higher than $\operatorname{cor}(R(\cdot), p)$; (ii) the near-miss trajectories receive rewards close to fully bad trajectories, whereas lucky guesses obtain rewards comparable to truly high-quality solutions. These results underscore the limitations of existing reward mechanisms, which remain dominated by outcome signals and pay insufficient attention to the effectiveness and robustness of reasoning patterns. More details and analyses are provided in \textbf{Appendix E.1}.

\begin{figure}
    \centering
    \includegraphics[width=0.9\linewidth]{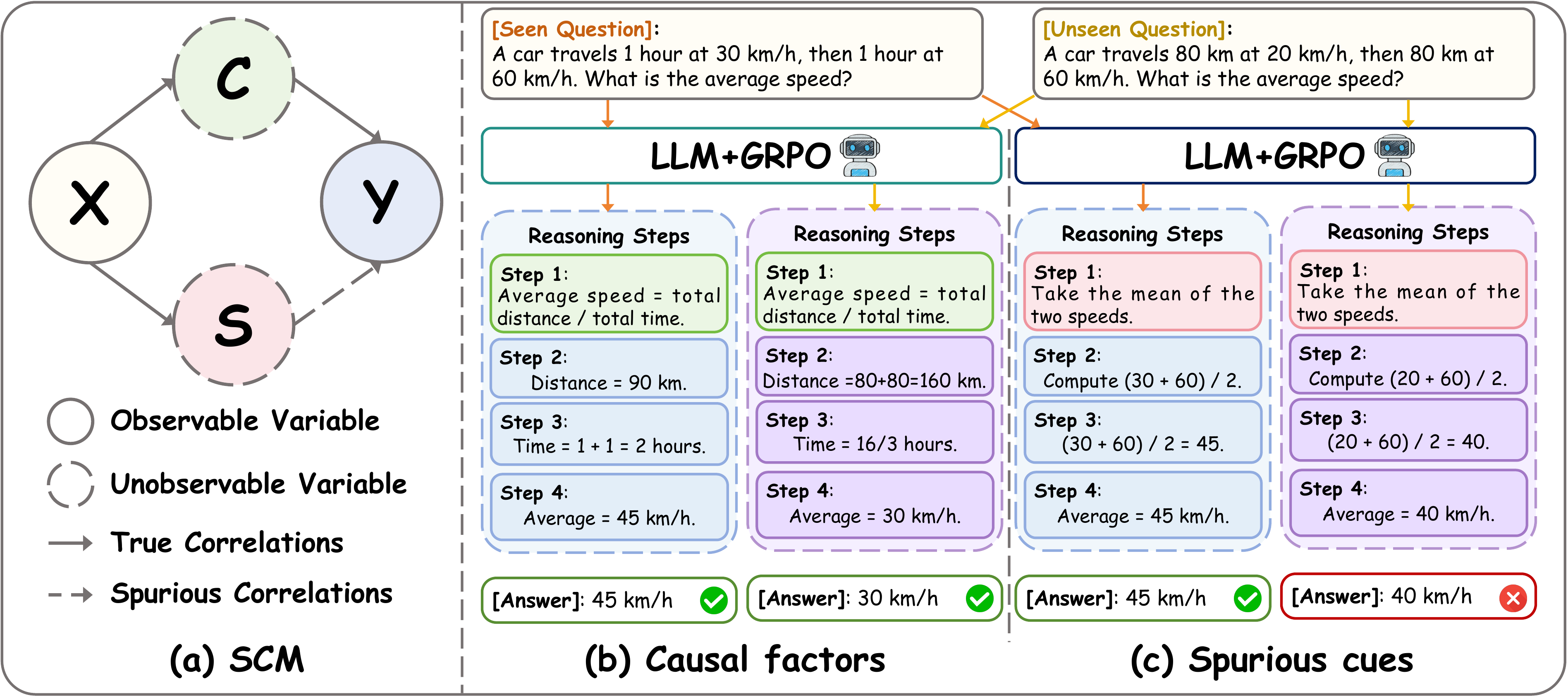}
    \caption{Motivation analyses. (a) shows the SCM of LLM reasoning. (b) and (c) provide examples illustrating how causal factors and spurious cues reinforced on seen questions fail to generalize.}
    \label{fig:scm_motivation}
    \vspace{-0.15in}
\end{figure}

\subsection{Motivation Analysis}
\label{sec:motivation_analysis}
To address this limitation, in this subsection, we discuss the key question: how can we design fine-grained rewards that decouple process validity from final correctness, so as to encourage the LLMs to learn generalizable reasoning strategies rather than merely mimicking answers.

We adopt a causal perspective and conduct a Structural Causal Model (SCM) for analyses (\textbf{Figure \ref{fig:scm_motivation}}). 
Following \citep{pearl2009causality}, we posit that for a given question $x$, the reasoning behavior of an LLM is jointly driven by two latent factors: (i) the causal factors $c$, which encodes task-governing mechanisms that remain valid under distribution shifts, e.g., arithmetic rules, logical consistency, and valid variable binding; and (ii) spurious cues $s$, which encapsulates brittle, dataset-specific shortcuts that coincide with correct answers only within a narrow context, e.g., reliance on specific formatting templates or shallow keyword associations. Achieving generalizable reasoning necessitates that the LLM relies on the invariant structure $c$ rather than overfitting to the spurious correlations $s$. However, a challenge arises from the observational equivalence of existing reward mechanisms: a shortcut driven by $s$ and a rigorous logical deduction driven by $c$ yield the same correct final answer $y$; the reward mechanisms that rely solely on verifying $y$ cannot distinguish between the two, frequently assigning equal credit to both of them. It may encourage shortcut behavior through $s\to y$. To mitigate this, we aim to design a reward that assigns higher scores with $c$ while filtering out $s$, encouraging the learning of generalizable reasoning. See \textbf{Appendix D.5} for details. 

To distinguish the causal factors $c$ from spurious cues $s$, we propose to leverage the sampling mechanism inherent in GRPO to approximate a causal analysis. 
Instead of evaluating a single trajectory in isolation, we view the generation of multiple candidates as a set of counterfactual experiments. Formally, we establish the theoretical support as follows:
\begin{theorem}
\label{theorem:motivation}
For a fixed input $x$, the reasoning of policy $\pi_\theta$ forms a Markov decision process (MDP) with transition kernel $P(s_{t+1}| s_t,a_t)$, satisfying: (i) There exists an SCM $\mathcal{M}$, without interventions, the trajectory distribution generated by $\mathcal{M}$ coincides with that of the MDP. (ii) For any alternative policy $\pi_\theta'$ or intervention on $P(s_{t+1}| s_t,a_t)$, the counterfactual trajectory distribution in $\mathcal{M}$ coincides with the distribution obtained by simulating the modified MDP.
\end{theorem}
\textbf{Theorem \ref{theorem:motivation}} is a specialization of the main results in \citep{oberst2019counterfactual,ness2019integrating}, which show that finite MDPs can be represented by SCMs where counterfactual distributions match the dynamics of the modified process. 
For a question $x$, the sampled trajectories in GRPO can be viewed as parallel counterfactual experiments under shared exogenous noise but different reasoning paths. This perspective shifts the focus from identifying a single successful outcome to evaluating the structural stability of the reasoning process itself. 
By analyzing the variations across these parallel candidates, we can identify reasoning nodes that maintain logical consistency under perturbations, distinguishing them from those that are brittle or accidental. See \textbf{Appendix B.1} for proofs.

Guided by this perspective, we establish two core principles for reward design: robustness and effectiveness. Robustness measures the stability of the answer distribution induced by a given reasoning step under local perturbations. A high robustness score indicates that the step relies on invariant logic $c$ rather than brittle spurious cues. However, robustness alone can favor degenerate or uninformative steps that remain trivially stable. We therefore introduce effectiveness. It measures the step's information content to ensure it conveys sufficient task-relevant information and prevents representation collapse. By integrating them, we move beyond binary outcome supervision. We reward sound reasoning even when the final answer is wrong, while penalizing trajectories that reach the right answer via flawed or spurious paths, thereby enforcing generalizable reasoning.


\section{Method}
\label{sec:method}

\begin{figure*}
    \centering
    \includegraphics[width=\linewidth]{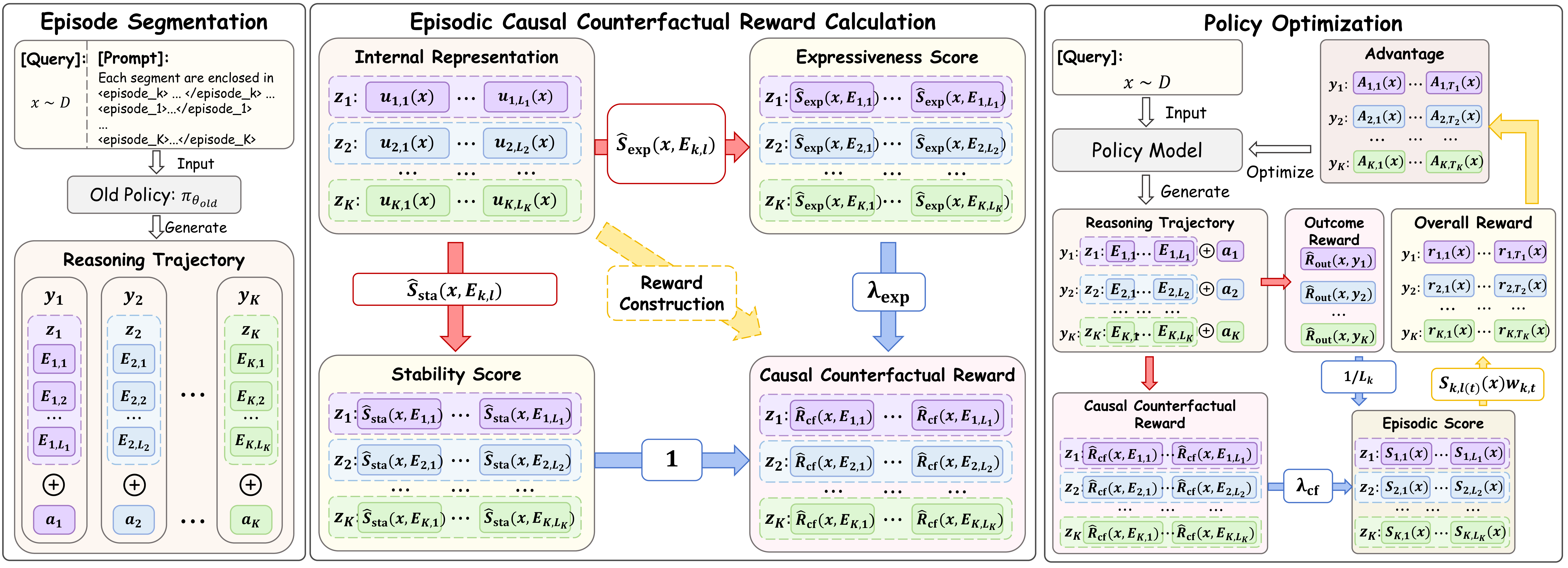}
    \caption{Illustration of G$C^2$PO. It segments reasoning into episodes (\textbf{Left}), calculates causal counterfactual reward (\textbf{Middle}), and optimizes LLMs by building token-level advantages (\textbf{Right}).}
    \label{fig:framework}
    \vspace{-0.1in}
\end{figure*}

Based on these analyses, we propose Group Causal Counterfactual Policy Optimization (GC$^2$PO). The core innovation is an episodic causal counterfactual reward. Instead of merely rewarding a trajectory that happens to be correct (which may rely on spurious shortcuts), it evaluates whether the underlying reasoning pattern is truly generalizable. GC$^2$PO operates in three key steps. First, we automatically segment each question-solution pair into distinct episodes via prompt design (\textbf{Subsection~\ref{sec:method_episode_segmentation}}). Then, we compute a causal counterfactual reward for each episode. This reward consists of two terms that jointly measure the robustness and effectiveness of the reasoning step (\textbf{Subsection~\ref{sec:method_reward}}). Finally, we leverage these rewards with the outcome reward to construct token-level advantages, optimizing the LLM to internalize generalizable reasoning strategies (\textbf{Subsection~\ref{sec:method_policy}}). The framework is shown in \textbf{Figure~\ref{fig:framework}}. The list of notations and pseudo-code are provided in \textbf{Appendix A} and \textbf{Appendix C}.

\subsection{Episode Segmentation}
\label{sec:method_episode_segmentation}

Reasoning tasks vary significantly in complexity. 
However, existing GRPO-based methods typically treat the entire solution as a single episode \citep{wiering2012reinforcement,szepesvari2022algorithms}. 
This coarse granularity is insufficient for complex reasoning: it ignores the internal logical structure and fails to identify which specific step caused a success or failure. 
To enable fine-grained evaluation, we reformulate the reasoning process as a sequence of distinct episodes, where each episode represents a coherent logical step.
Formally, given a query $x \sim \mathcal{D}$, the policy $\pi_{\theta_\mathrm{old}}$ generates a group of $K$ candidate trajectories $\{y_k\}_{k=1}^K$. 
Each trajectory $y_k$ consists of a reasoning chain $z_k$ and a final answer $a_k$. 
We decompose the reasoning chain $z_k$ into $L_k$ discrete episodes, denoted as $z_k = \{E_{k,l}\}_{l=1}^{L_k}$, where each $E_{k,l}$ corresponds to the tokens of the $l$-th reasoning step. In practice, we enforce this segmentation via prompt engineering. 
The model is instructed to delimit reasoning steps using specific tags (e.g., \texttt{<episode\_i>} and \texttt{</episode\_i>}, see \textbf{Appendix E.5}), ensuring that each $E_{k,l}$ forms a self-contained semantic unit. 
It allows us to assign rewards to individual logical steps, enabling the model to learn from correct intermediate steps even if the final answer is incorrect.

\subsection{Causal Counterfactual Reward Calculation}
\label{sec:method_reward}
To determine whether an episode encodes a generalizable reasoning pattern, we propose an episodic causal counterfactual reward.
The reward consists of: (i) a \emph{stability term}: for robustness, it measures how much of the reasoning behavior of the episode survives under latent perturbations, and (ii) an \emph{expressiveness term}: for effectiveness, it constrains the decay rate of the episode's representation, ensuring that the reasoning step retains sufficient information to be effective across different contexts. 
By optimizing for both, the LLM is incentivized to converge toward reasoning patterns that are both robust and effective.
In this subsection, we first clarify the roles of the two terms and then formally define the reward (\textbf{Theorem~\ref{theorem:definition_reward}}).
Next, we show how to compute it efficiently with theoretical guarantees (\textbf{Theorem~\ref{theorem:reward}}). Finally, we discuss why this reward is effective.

To simulate potential variations in the reasoning context, we operate within the continuous representation space. It offers a more stable and computationally efficient alternative to perturbing discrete input tokens or model parameters \citep{wang2025learning,wang2024moreaupruner}.
For a given question $x$ and episode $E_{k,l}$, we map the reasoning step to a latent vector $u_{k,l}(x, E_{k,<l} \in \mathbb{R}^d$ (simpled write as $u_{k,l}(x)$), which is derived from the hidden state of the last token. This vector determines the model's answer distribution $q(\cdot | u_{k,l}(x))$ over the answer space $\mathcal{A}$.
We then apply a family of perturbation operators $\{g_m\}_{m=1}^M$ to generate perturbed representations $\tilde u_{k,l}^{(m)}(x) = g_m(u_{k,l}(x))$ and their induced answer distributions $q(\cdot | \tilde u_{k,l}^{(m)}(x))$.
Intuitively, $u_{k,l}(x)$ encodes ``what this episode is trying to do'', while $\tilde u_{k,l}^{(m)}(x)$ simulates nearby ``what-if'' scenarios. 
The stability term quantifies the divergence between $q(\cdot | u_{k,l}(x))$ and $q(\cdot | \tilde u_{k,l}^{(m)}(x))$: episodes whose logic truly works under variations keep $q(\cdot | u_{k,l}(x))$ and $q(\cdot | \tilde u_{k,l}^{(m)}(x))$ close and thus receive high scores; brittle episodes exhibit large shifts and are penalized. The \emph{expressiveness term} is defined from the viewpoint of the optimal decay rate \citep{cazenave1998introduction,pazy2012semigroups}: it tracks how fast the representation $u_{k,l}(x)$ decays along perturbation directions. If the perturbations quickly drive $u_{k,l}(x)$ toward a low-norm, uninformative state, the episode is structurally fragile and gets a low score; if the representation remains strong and decays slowly, the episode is rewarded. 
Together, they encourage episode representations that are both stable in their predictions and persistent in representation space. 

We then present the formal definition of the episodic causal counterfactual reward with causal identifiability guarantees as follows. More analyses are provided in \textbf{Appendix D.3}.
\begin{theorem}
\label{theorem:definition_reward}
For a fixed $x$ and episode $E_{k,l}$, let $u_{k,l}(x)\in\mathbb{R}^d$ denote the episode representation and $q(\cdot | u_{k,l}(x))\in\Delta(\mathcal{A})$ the induced answer distribution. Let $\{g_m\}_{m=1}^M$ be a family of perturbation operators acting on representations, and define \scalebox{0.95}{$\tilde u_{k,l}^{(m)}(x) = g_m(u_{k,l}(x))$} and \scalebox{0.95}{$q^{(m)}(\cdot | x) = q(\cdot | \tilde u_{k,l}^{(m)}(x))$}. Let $\pi^*_\theta$ be any policy that achieves the highest causal counterfactual reward $R_{\mathrm{cf}}^{\star}(x,E_{k,l})$:
\begin{equation}
\label{eq:definition_reward}
\begin{gathered}
R_{\mathrm{cf}}^{\star}(x,E_{k,l})=S_{\mathrm{sta}}(x,E_{k,l})+\lambda_{\mathrm{exp}}\,S_{\mathrm{exp}}(x,E_{k,l}),\\
\text{s.t.}\;\scalebox{0.95}{$S_{\mathrm{sta}}(x,E_{k,l})=\mathbb{E}_{m}\Big[\exp\Big(-\tfrac{1}{\tau}\big\|q(\cdot | u_{k,l}(x))-q^{(m)}(\cdot | x)\big\|_2^2\Big)\Big],$}\;
\scalebox{0.95}{$S_{\mathrm{exp}}(x,E_{k,l})=
\mathbb{E}_{m}\Bigg[\frac{\big\|\tilde u_{k,l}^{(m)}(x)\big\|_2^2}{\big\|u_{k,l}(x)\big\|_2^2 + \varepsilon_u}\Bigg],$}
\end{gathered}
\end{equation}
where $\lambda_{\mathrm{exp}}$ is weighting coefficient, $\tau$ and $\varepsilon_u$ are hyperparameters following \citep{pazy2012semigroups}.
Then, $\pi^*_\theta$ block-identifies the true causal variables in the sense of \textbf{Definition B.3}.
\end{theorem}

\emph{\textbf{Intuition and Discussion.}} A larger $R_{\mathrm{cf}}^{\star}$ indicates a more valuable episode for optimization. $S_{\mathrm{sta}}$ is maximized when $q(\cdot | u_{k,l}(x))=q^{(m)}(\cdot | x)$ for all $m$, and decays exponentially as these distributions diverge; it therefore measures distributional robustness under pertubations. $S_{\mathrm{exp}}(x,E_{k,l})$ captures how well the representation avoids collapsing under perturbations: with a local linearization $ \tilde u_{k,l}^{(m)}(x)\approx A_m u_{k,l}(x)$, the ratio estimates the squared gain of $A_m$ along $u_{k,l}(x)$, and its expectation reflects the effective energy decay. Larger $S_{\mathrm{exp}}$ thus implies weaker decay and more informative representations. 
Together, these two terms guarantee that the learned representation is invariant to nuisances while fully retaining essential causal information. By the inverse mapping theorem, this establishes an invertible correspondence between $u_{k,l}(x)$ and the underlying true causal variables, formally achieving block-identifiability.
See \textbf{Appendix B.2} for more details.

Obtaining \textbf{Theorem~\ref{theorem:definition_reward}}, we illustrate how to calculate it in practice.
For each $x$, we run a forward pass to obtain $u_{k,l}(x)$ and $q(\cdot | u_{k,l}(x))$ with \textbf{Subsection \ref{sec:method_episode_segmentation}}, then sample $M$ perturbed representations $\tilde u_{k,l}^{(m)}(x)$. 
The implementation of the latent-space perturbation operators $\{g_m\}_{m=1}^M$ follows \citep{pearl2009causality,kusner2017counterfactual}, satisfying the minimal change principle for counterfactuals. More specifically, for the final boundary token of the $l$-th reasoning episode in trajectory $y_k$, we extract its hidden state vector $u_{k,l}(x) \in \mathbb{R}^d$ after the final normalization layer. The perturbation operator $g_m: \mathbb{R}^d \to \mathbb{R}^d$ is defined as an adaptive isotropic Gaussian diffusion in the latent space (whose tangential components excite local perturbations on the manifold), i.e., \scalebox{0.92}{$\tilde{u}_{k,l}^{(m)}(x) = g_m(u_{k,l}(x)) = u_{k,l}(x) + \delta_m$}, where the standard deviation is dynamically scaled according to the norm of the current representation: \scalebox{0.92}{$\sigma = \alpha \cdot \frac{\Vert{}u_{k,l}(x)\Vert{}_2}{\sqrt{d}}$}. This adaptive scaling confines the perturbation within an infinitesimal first-order differential neighborhood of the factual state.
Next, we compute the answer distributions $q^{(m)}(\cdot | x)$ and norms, and substitute them into Eq.\ref{eq:definition_reward} for $S_{\mathrm{sta}}$ and $S_{\mathrm{exp}}$.
Since computing $S_{\mathrm{sta}}(x,E_{k,l})$ and $S_{\mathrm{exp}}(x,E_{k,l})$ requires expectations over the full perturbation distribution, which are intractable in closed form, we approximate them with Monte Carlo estimators on $M$ samples.
Below, we establish:
\begin{theorem}
\label{theorem:reward}
Let the latent manifold be $\mathcal{Z} = \mathcal{C} \times \mathcal{S}$ endowed with a Riemannian metric, assume the perturbations $\{g_m\}_{m=1}^M$ locally span the tangent space of the non-causal manifold $T_s\mathcal{S}$.
If a learned $u^*$ satisfies $S_{\mathrm{sta}} \ge 1 - \epsilon$ and $S_{\mathrm{exp}} \ge \gamma$, then, maximizing $R_{\mathrm{cf}}^{\star}$ minimizes the error bound to the oracle policy $\pi_{\mathrm{opt}}(\cdot|c) := \mathbb{E}_{s}[\pi^*(\cdot|c,s)]$ with $\mathbb{E}_s[D_{\mathrm{KL}}(\pi_{\mathrm{opt}}(\cdot|c) \,\|\, \pi^*(\cdot|c, s))] \le \mathcal{O}(\sqrt{\epsilon})$.
\end{theorem}
\textbf{Theorem \ref{theorem:reward}} establishes that maximizing our proposed reward drives the learned policy to converge toward the optimal causal oracle $\pi_{\mathrm{opt}}$. See \textbf{Appendix B.3} for proofs.

Finally, we explain why the reward is effective. It operates at the episode level and evaluates how the encoded reasoning generalizes, rather than only final correctness. The episodes containing genuinely valid reasoning steps tend to induce answer distributions that remain stable under mechanism-preserving perturbations, yielding high stability scores $S_{\mathrm{sta}}(x,E_{k,l})$; while their representations do not collapse or rapidly decay along perturbation directions, leading to large expressiveness scores $S_{\mathrm{exp}}(x,E_{k,l})$. In contrast, brittle shortcuts or filler steps produce large distribution shifts or rapid representation decay, leading to low $R_{\mathrm{cf}}(x,E_{k,l})$ even when the final answer is accidentally correct. Thus, the reward provides a fine-grained signal that separates process validity from outcome correctness and steers the model toward robust, reusable reasoning patterns.

\subsection{Policy Optimization}
\label{sec:method_policy}
In this subsection, we illustrate the policy optimization of G$C^2$PO.
Specifically, we integrate the proposed causal counterfactual reward with the outcome reward in a GRPO-style objective, considering both reasoning effectiveness and answer correctness. For each candidate $y_k$ and episode index $l$, we define an episodic score that combines both rewards:
\begin{equation}
S_{k,l}(x)=\frac{1}{L_k} R_{\mathrm{out}}(x,y_k)+\lambda_{\mathrm{cf}}R_{\mathrm{cf}}(x,E_{k,l}),
\end{equation}
where $L_k$ is the total number of reasoning episodes in trajectory $y_k$, $R_{\mathrm{cf}}(x,E_{k,l})$ is the causal counterfactual reward of episode $E_{k,l}$, $\lambda_{\mathrm{cf}}$ is a balancing hyperparameter, and \scalebox{0.92}{$R_{\mathrm{out}}(x,y_k)\in\{0,1\}$} is the binary outcome reward evenly distributed across episodes.
We then distribute episodic scores to tokens. For the $t$-th token of trajectory $k$, let $l$ be the episode index such that $y_{k,t}\in E_{k,l}$. We define an unnormalized surprise weight under the old policy, i.e., \scalebox{0.95}{$\tilde w_{k,t}(x)=-\log \pi_{\theta_\mathrm{old}}\big(y_{k,t}\mid x, y_{k,<t}\big)$},
and normalize it as $w_{k,t}(x)$.
The resulting token-level reward $r_{k,t}(x)$ is:
\begin{equation}\label{eq:reward_token}
\begin{aligned}
r_{k,t}(x)=S_{k,l(t)}(x)\,w_{k,t}(x),\quad\text{s.t.}\;w_{k,t}(x)=\frac{\tilde w_{k,t}(x)}{\sum_{t'\in E_{k,l(t)}} \tilde w_{k,t'}(x)}.
\end{aligned}
\end{equation}
For each trajectory $k$, we aggregate the token-level rewards into a single scalar by taking a truncated mean over tokens (i.e., discarding a small fraction of extreme values), denoted by $\tilde r_k(x)=\operatorname{TruncMean}\big(\{r_{k,t}(x)\}_{t=1}^{T_k}\big)$.
Across the $K$ candidates for the same $x$, we compute the group-normalized advantage $\hat A_k(x)$ with group-wise mean $\bar r(x)$ and variance $s_r^2(x)$ of these trajectory-level scores. We rescale this advantage back to tokens according to their relative contribution:
\begin{equation}
\begin{aligned}
    & \scalebox{0.98}{$A_{k,t}(x)=\hat A_k(x)\,\frac{r_{k,t}(x)}{\tilde r_k(x)},\;\text{s.t.}\;
    \hat A_k(x)=\frac{\tilde r_k(x) - \bar r(x)}{\sqrt{s_r^2(x)}},$}\\
    &\scalebox{0.98}{$
    \bar r(x)=\frac{1}{K}\sum_{k=1}^K \tilde r_k(x),\;s_r^2(x)=\frac{1}{K}\sum_{k=1}^K\big(\tilde r_k(x) - \bar r(x)\big)^2.$}
\end{aligned}
\end{equation}
In this way, even trajectories with $R_{\mathrm{out}}(x,y_k)=0$ can receive positive advantages on episodes with high counterfactual rewards, converting previously negative groupwise advantages into useful learning signals for genuinely good but unlucky reasoning. Conversely, for trajectories with $R_{\mathrm{out}}(x,y_k)=1$, useless episodes obtain relatively small or even negative advantages after normalization, discouraging the model from reinforcing unstable reasoning patterns.

G$C^2$PO then adopts a GRPO-style optimization objective at the token level: 
\begin{equation}
\label{eq:G$C^2$PO_total}
\begin{aligned}
&\mathcal{J}_{\mathrm{GC^2PO}}(\theta)=
\mathbb{E}_{\,x \sim \mathcal{D},\, \{y_k\}\sim \pi_{\theta_\mathrm{old}}(\cdot\mid x)}\Big[\frac{1}{K}\sum_{k=1}^K \frac{1}{T_k}\sum_{t=1}^{T_k}g_{k,t}(\theta,x)-\beta_{\mathrm{KL}}\,
\mu_{\mathrm{KL}}\big(\pi_\theta\,\Vert\,\pi_{\mathrm{ref}}\big)
\Big],
\\
&\text{s.t.}\;g_{k,t}(\theta,x)=\min\Big(\rho_{k,t}(\theta)\,A_{k,t}(x),\;
\tilde\rho_{k,t}(\theta)\,A_{k,t}(x)\Big),
\end{aligned}
\end{equation}
where $\beta_{\mathrm{KL}}$, $\rho_{k,t}(\theta)$, and $\tilde\rho_{k,t}(\theta)$ are the same as in Eq.\ref{eq:problem_grpo}.
Based on this, G$C^2$PO drives the LLMs to generate episodes that are both outcome-successful and counterfactually robust, thereby learning effective and generalizable reasoning strategies across diverse questions and tasks.

\begin{table*}[t]
  \centering
   \caption{Pass@1 performance on mathematical reasoning datasets. We compare base models trained with different approaches. The best results are highlighted in \textbf{bold}. ``()'' denotes the improvement over models without fine-tuning. More results are provided in \textbf{Appendix E}.}
  \label{tab:ex_1}
  \small
  \resizebox{\linewidth}{!}{%
    \begin{tabular}{l|c|c|c|c|c|c|c}
      \toprule
      \textbf{Methods} & \textbf{AIME 2024} & \textbf{AIME 2025} & \textbf{AMC 2023} & \textbf{MATH500} & \textbf{MinervaMATH} & \textbf{GSM8K} & \textbf{Average Results} \\
      \midrule
      \!\textbf{DeepScaleR‑1.5B‑Preview} & 42.8 & 36.7 & 83.0 & 85.2 & 24.6 & 89.6 & 60.3 \\
      \!~~~~+ GRPO \citep{shao2024deepseekmath} & 44.5 (+1.7) & 39.3 (+2.6) & 81.5 (-1.5) & 84.9 (-0.3) & 24.7 (+0.1) & 89.4 (-0.2) & 60.7 (+0.4) \\
      \!~~~~+ length penalty \citep{arora2025training} & 40.3 (-2.5) & 30.3 (-6.4) & 77.3 (-5.7) & 83.2 (-2.0) & 23.0 (-1.6) & 88.5 (-1.1) & 57.1 (-3.2) \\
      \!~~~~+ ReST-MCTS \citep{zhang2024rest}& 45.5 (+2.7) & 39.5 (+2.8) & 83.4 (+0.4) & 84.8 (-0.4) & 23.9 (-0.7) & 89.9 (+0.3) & 61.2 (+0.9) \\
      \!~~~~+ GVPO \citep{zhang2025gvpo} & 46.1 (+3.3) & 39.7 (+3.0) & 83.6 (+0.6) & 85.7 (+0.5) & 25.3 (+0.7) & 90.4 (+0.8) & 61.8 (+1.5) \\
      \!~~~~+ Dr.GRPO \citep{liu2025understanding} & 45.8 (+3.0) & 39.6 (+2.9) & 82.1 (-0.9) & 85.3 (+0.1) & 25.1 (+0.5) & 90.0 (+0.4) & 61.3 (+1.0) \\
       \!~~~~+ GCPO \citep{gu2025group} & 46.7 (+3.9) & 40.3 (+3.6) & 84.1 (+1.1) & 86.3 (+1.1) & 25.9 (+1.4) & 90.5 (+0.9) & 62.3 (+2.0) \\
       \!~~~~+ MRT \citep{qu2025optimizing} & 47.2 (+4.4) & 39.7 (+3.0) & 83.1 (+0.1) & 85.1 (-0.1) & 24.2 (-0.4) & 89.9 (+0.3) & 61.5 (+1.2) \\
       \rowcolor{orange!10} \!~~~~+ G$C^2$PO (Ours) & \textbf{49.3 (+6.2)} & \textbf{40.6 (+3.9)} & \textbf{86.1 (+3.1)} & \textbf{88.3 (+3.1)} & \textbf{27.5 (+2.9)} & \textbf{91.6 (+2.0)} & \textbf{63.9 (+3.6)} \\

      \midrule
      \!\textbf{DeepSeek-R1-Distill-Qwen-1.5B} & 28.7 & 26.0 & 69.9 & 80.1 & 19.8 & 83.4 & 51.3 \\
      \!~~~~+ GRPO \citep{shao2024deepseekmath} & 29.8 (+1.1) & 27.3 (+1.3) & 70.5 (+0.6) & 80.3 (+0.2) & 22.1 (+2.3) & 84.5 (+1.1) & 52.4 (+1.1) \\
      \!~~~~+ length penalty \citep{arora2025training} & 27.5 (-1.2) & 22.6 (-3.4) & 64.4 (-5.5) & 77.1 (-3.0) & 18.8 (-1.0) & 82.6 (-0.8) & 48.8 (-2.5) \\
      \!~~~~+ ReST-MCTS \citep{zhang2024rest} & 30.5 (+1.8) & 28.6 (+2.6) & 71.1 (+1.2) & 80.4 (+0.3) & 20.3 (+0.5) & 84.8 (+1.4) & 52.6 (+1.3) \\
      \!~~~~+ GVPO \citep{zhang2025gvpo} & 30.6 (+1.9) & 28.2 (+2.2) & 71.5 (+1.6) & 80.5 (+0.4) & 23.1 (+3.3) & 85.0 (+1.6) & 53.2 (+1.8) \\
      \!~~~~+ Dr.GRPO \citep{liu2025understanding} & 30.4 (+1.7) & 28.4 (+2.4) & 71.3 (+1.4) & 80.8 (+0.7) & 22.9 (+3.1) & 85.0 (+1.6) & 53.1 (+1.8) \\
      \!~~~~+ GCPO \citep{gu2025group} & 31.0 (+2.3) & 29.0 (+3.0) & 71.8 (+1.9) & 81.6 (+1.5) & 23.4 (+3.6) & 85.3 (+1.9) & 53.7 (+2.4) \\
      \!~~~~+ MRT \citep{qu2025optimizing} & 30.3 (+1.6) & 29.3 (+3.3) & 72.9 (+3.0) & 80.4 (+0.3) & 22.5 (+2.7) & 84.7 (+1.3) & 53.4 (+2.0) \\
    \rowcolor{orange!10} \!~~~~+ G$C^2$PO (Ours) & \textbf{33.8 (+5.1)} & \textbf{30.7 (+4.7)} & \textbf{74.3 (+4.4)} & \textbf{85.3 (+5.2)} & \textbf{25.4 (+5.7)} & \textbf{86.5 (+3.1)} & \textbf{55.9 (+4.6)} \\ 
      
         \midrule
         
    \!\textbf{DeepSeek-R1-Distill-Qwen-7B} & 55.5 & 50.2 & 85.1 & 87.4 & 42.1 & 91.6 & 68.6 \\
    \!~~~~+ GRPO \citep{shao2024deepseekmath} & 56.9 (+1.4) & 51.7 (+1.5) & 85.5 (+0.4) & 87.7 (+0.3) & 43.5 (+1.4) & 92.1 (+0.5) & 69.6 (+0.9) \\
    \!~~~~+ length penalty \citep{arora2025training} & 53.8 (-1.7) & 46.9 (-3.3) & 81.2 (-3.9) & 83.7 (-3.7) & 39.5 (-2.6) & 91.1 (-0.5) & 66.0 (-2.6) \\
    \!~~~~+ ReST-MCTS \citep{zhang2024rest}  & 57.1 (+1.6) & 52.4 (+2.2) & 85.7 (+0.6) & 87.9 (+0.5) & 42.8 (+0.7) & 92.0 (+0.4) & 69.7 (+1.0) \\
    \!~~~~+ GVPO \citep{zhang2025gvpo}       & 57.5 (+2.0) & 52.1 (+1.9) & 86.3 (+1.2) & 88.5 (+1.1) & 44.2 (+2.1) & 92.9 (+1.3) & 70.3 (+1.6) \\
    \!~~~~+ Dr.GRPO \citep{liu2025understanding} & 57.4 (+1.9) & 52.3 (+2.1) & 86.4 (+1.3) & 88.2 (+0.8) & 44.0 (+1.9) & 92.3 (+0.7) & 70.1 (+1.5) \\
    \!~~~~+ GCPO \citep{gu2025group} & 58.3 (+2.8) & 53.0 (+2.8) & 87.3 (+2.2) & 89.1 (+1.7) & 45.0 (+2.9) & 92.6 (+1.0) & 70.9 (+2.2) \\
    \!~~~~+ MRT \citep{qu2025optimizing} & 57.0 (+1.5) & 52.4 (+2.2) & 86.0 (+0.9) & 88.4 (+1.0) & 44.3 (+2.2) & 92.2 (+0.6) & 70.1 (+1.4) \\
    \rowcolor{orange!10} \!~~~~+ G$C^2$PO (Ours) & \textbf{59.1 (+3.6)} & \textbf{54.3 (+4.1)} & \textbf{88.2 (+3.1)} & \textbf{89.8 (+2.4)} & \textbf{45.6 (+3.5)} & \textbf{93.6 (+2.0)} & \textbf{71.8 (+3.1)} \\
      \bottomrule
    \end{tabular}%
  }
\end{table*}

\begin{figure*}[t]
\centering
\begin{minipage}{0.49\textwidth} 
    \centering 
    
    \begin{subfigure}[b]{0.49\textwidth}
    \centering
    \includegraphics[width=\textwidth]{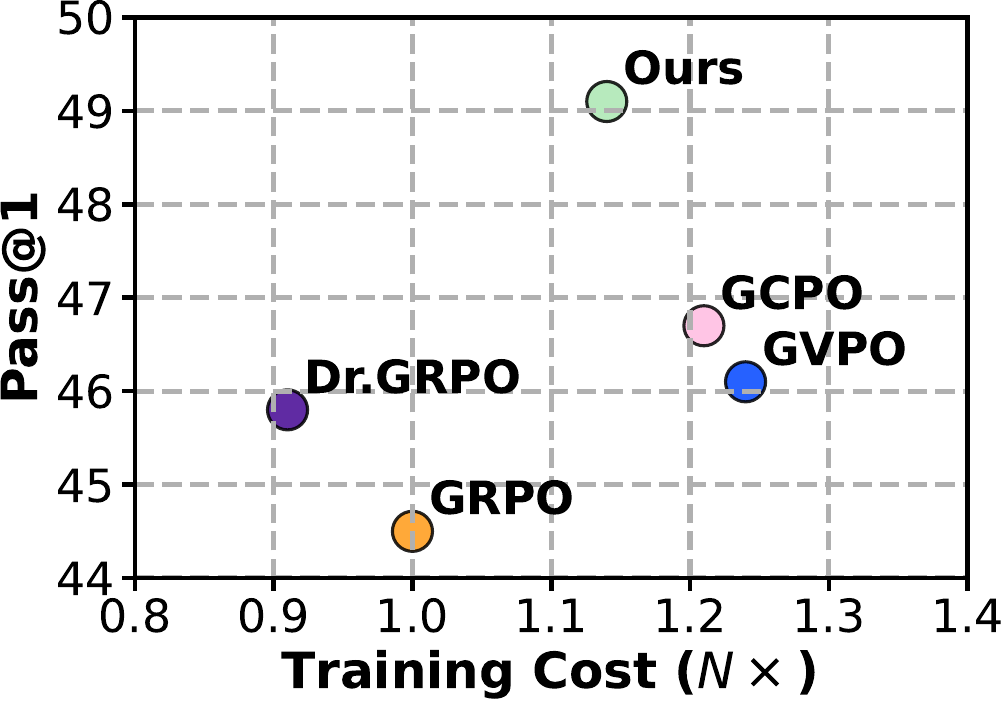}
    \caption{Training time.}
    \vspace{-0.05in}
    \label{fig:efficiency_cost_1}
  \end{subfigure}
  \hfill
  \begin{subfigure}[b]{0.49\textwidth}
    \centering
    \includegraphics[width=\textwidth]{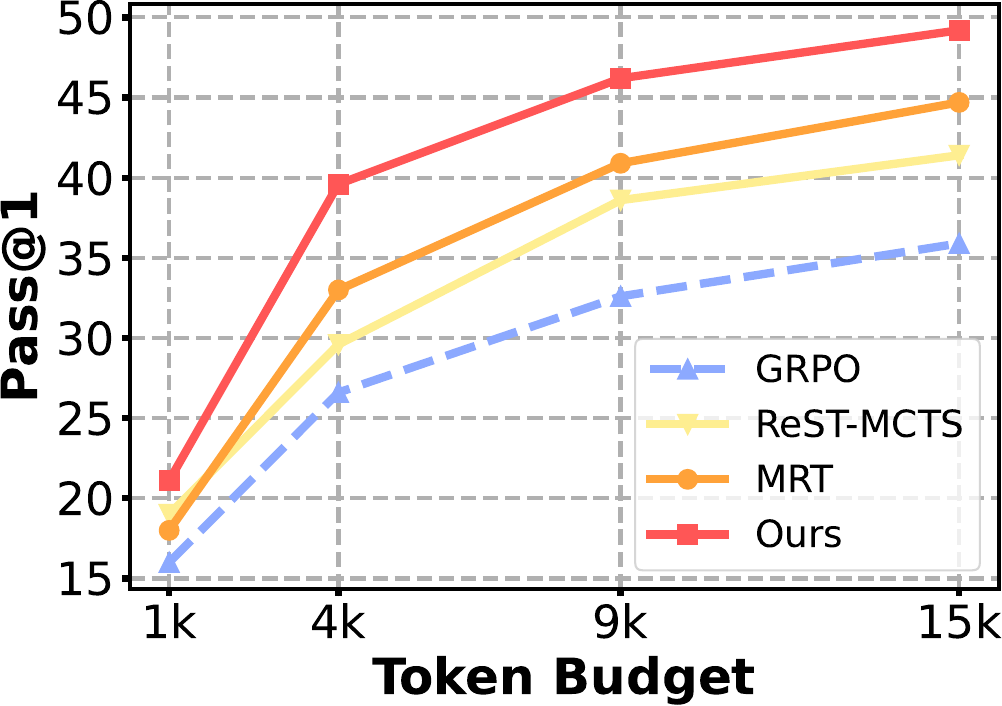}
    \caption{Token budget.}
    \vspace{-0.05in}
    \label{fig:efficiency_cost_2}
  \end{subfigure}
  
    \caption{Trade-off performance analyses.}
    \label{fig:efficiency_cost}
    \vspace{-0.1in}
\end{minipage}
\hfill 
\begin{minipage}{0.49\textwidth} 
    \centering 
    
  \begin{subfigure}[b]{0.49\textwidth}
    \centering
    \includegraphics[width=\textwidth]{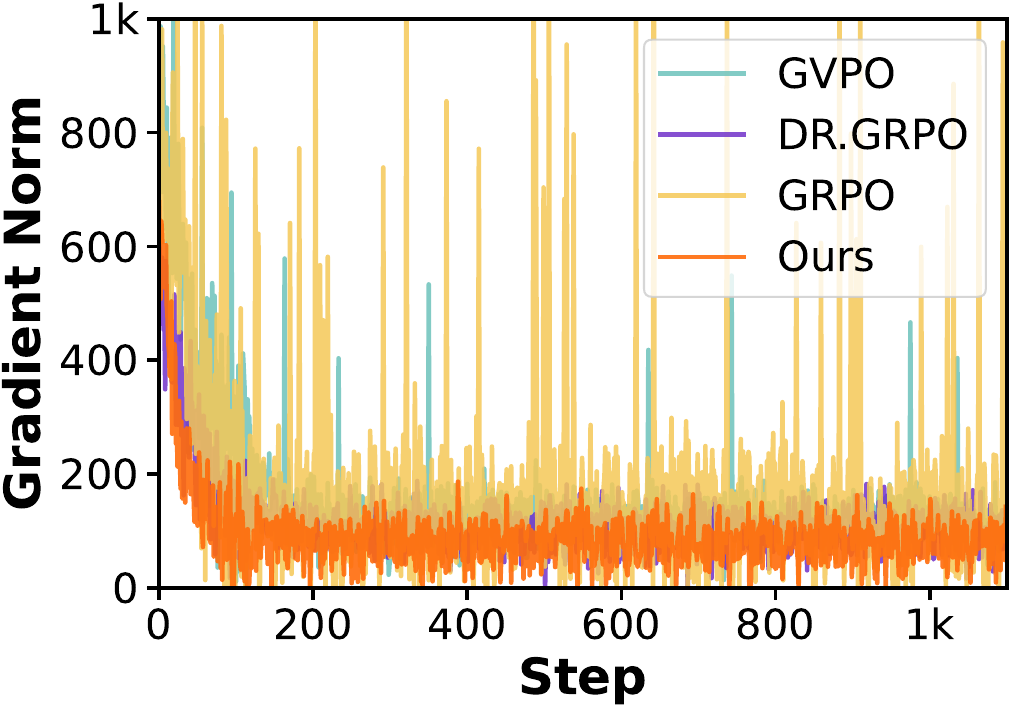}
    \caption{Gradient norm.}
    \vspace{-0.05in}
    \label{fig:norm_a}
  \end{subfigure}
  \hfill
  \begin{subfigure}[b]{0.49\textwidth}
    \centering
    \includegraphics[width=\textwidth]{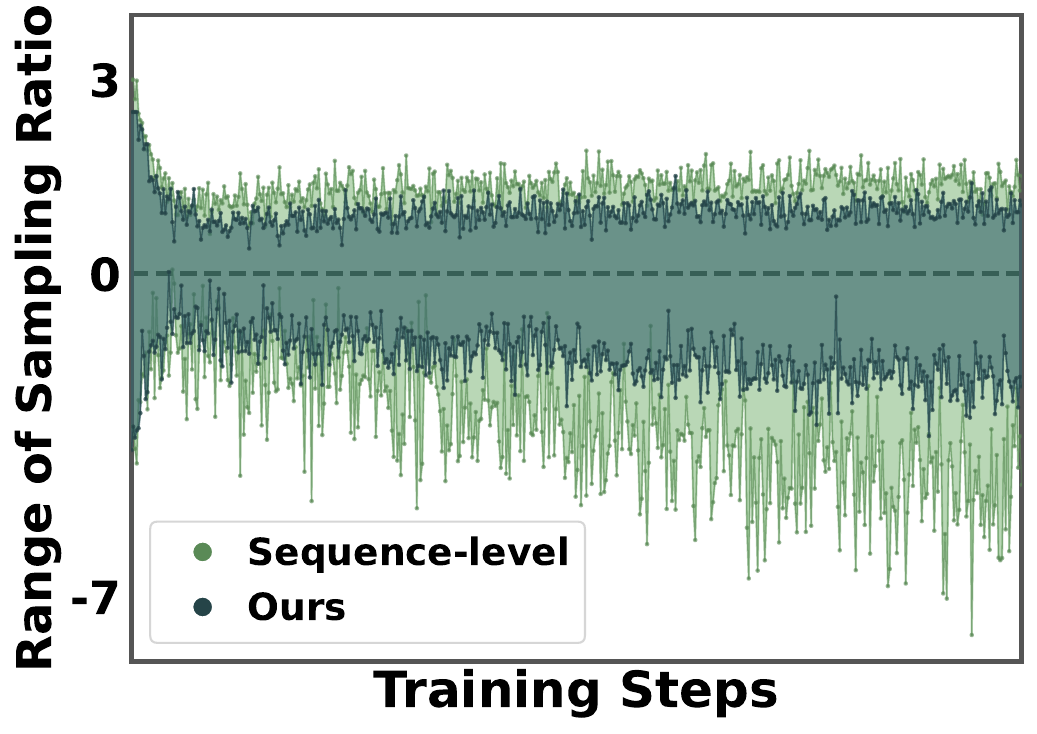}
    \caption{Sampling ratio.}
    \vspace{-0.05in}
    \label{fig:norm_b}
  \end{subfigure}
  \caption{Evaluation of training stability.}
  \label{fig:norm}
  \vspace{-0.1in}
\end{minipage}
\end{figure*}

\begin{figure}[t]
    \centering
        
    \begin{subfigure}[b]{0.22\textwidth}
    \centering
    \includegraphics[width=\textwidth]{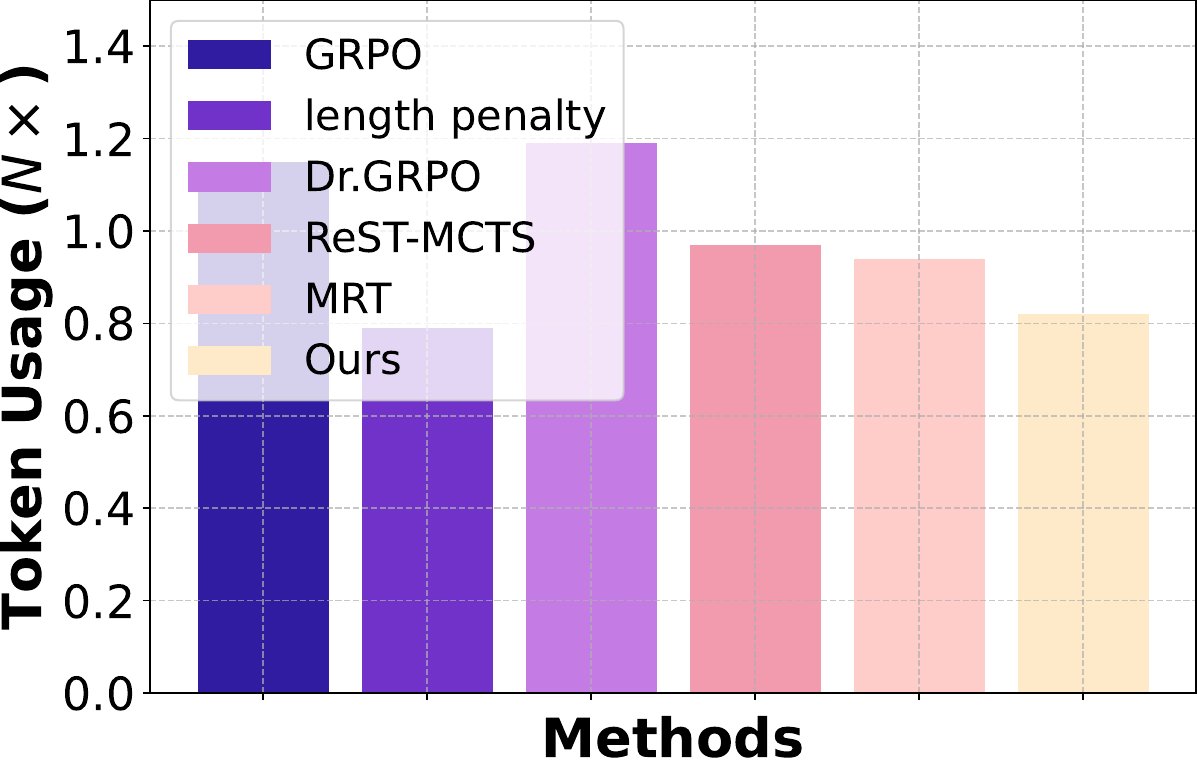}
    \caption{AIME}
  \end{subfigure}
  \hfill
  \begin{subfigure}[b]{0.22\textwidth}
    \centering
    \includegraphics[width=\textwidth]{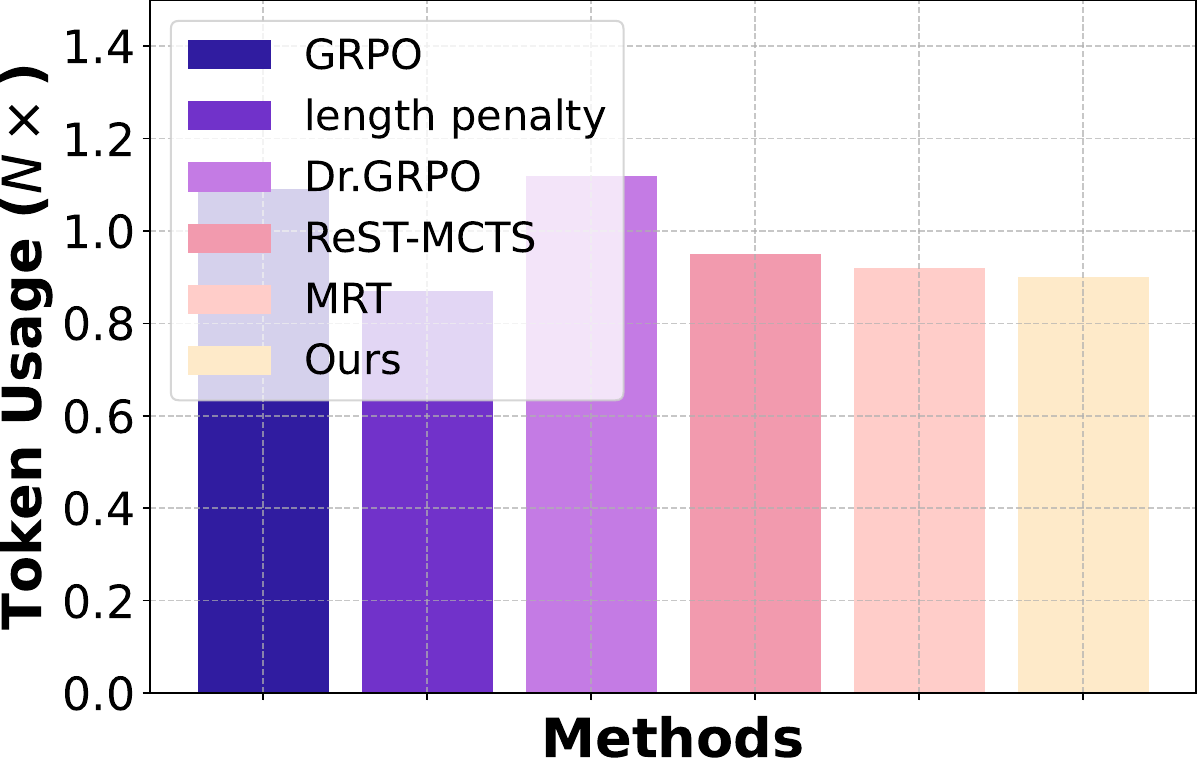}
    \caption{AMC}
  \end{subfigure}
  \hfill
  \begin{subfigure}[b]{0.22\textwidth}
    \centering
    \includegraphics[width=\textwidth]{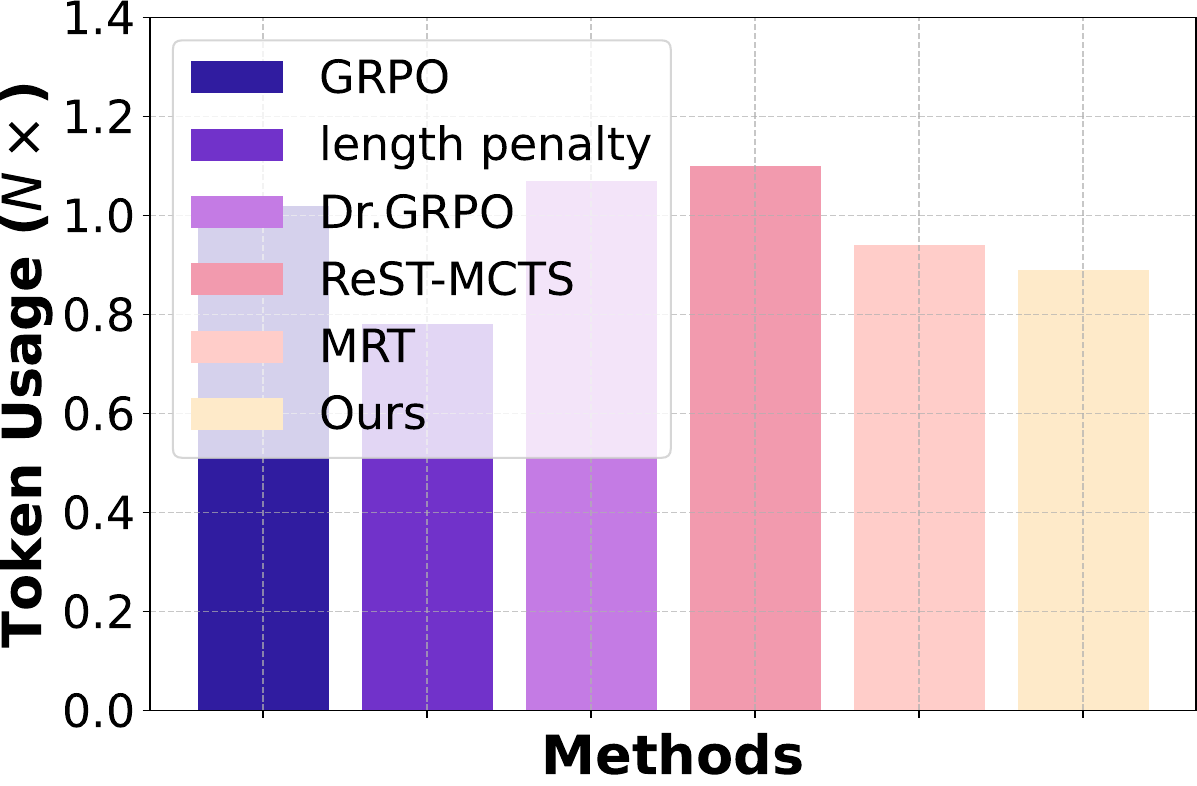}
    \caption{MATH500}
  \end{subfigure}
  \hfill
  \begin{subfigure}[b]{0.22\textwidth}
    \centering
    \includegraphics[width=\textwidth]{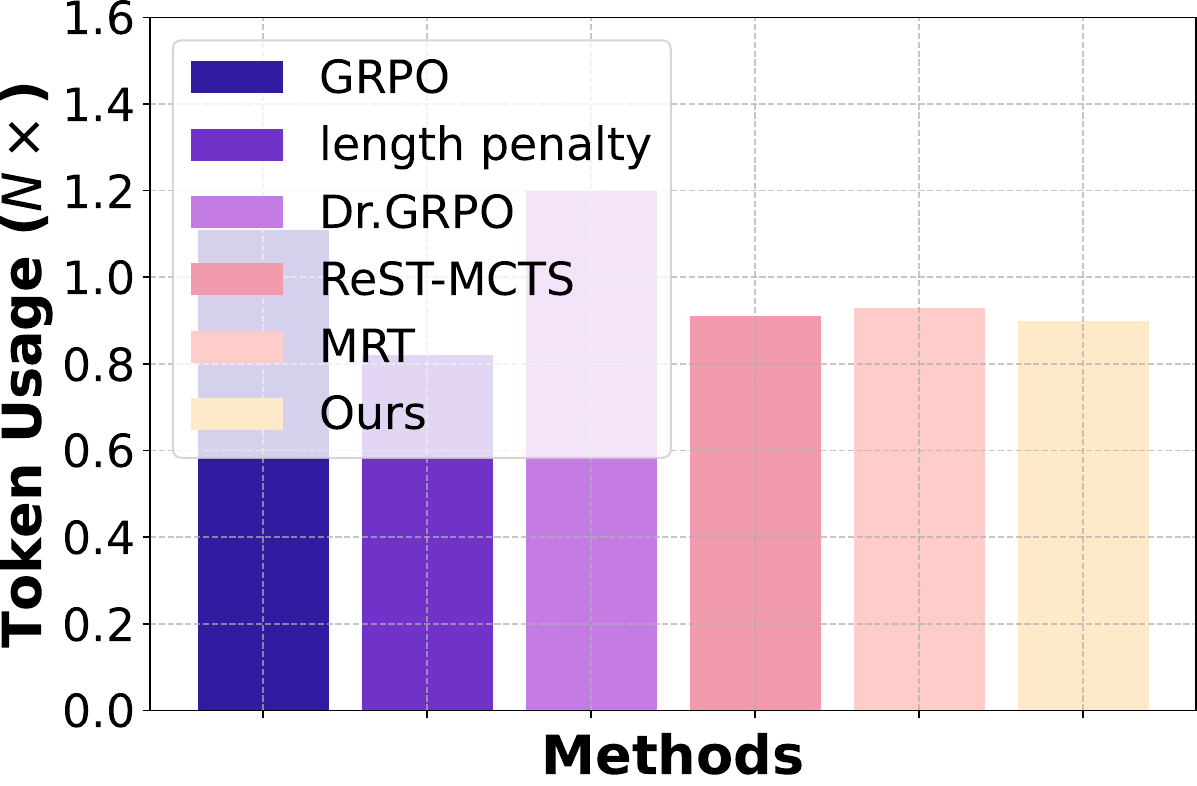}
    \caption{MinervaMATH}
  \end{subfigure}
    \caption{Comparison of token usage across different post-training methods.}
    \label{fig_app:ex_token_usage}
    \vspace{-0.1in}
\end{figure}

\begin{figure}[t]
    \centering
    \begin{minipage}[b]{0.49\linewidth}
        \centering
        \includegraphics[width=\linewidth]{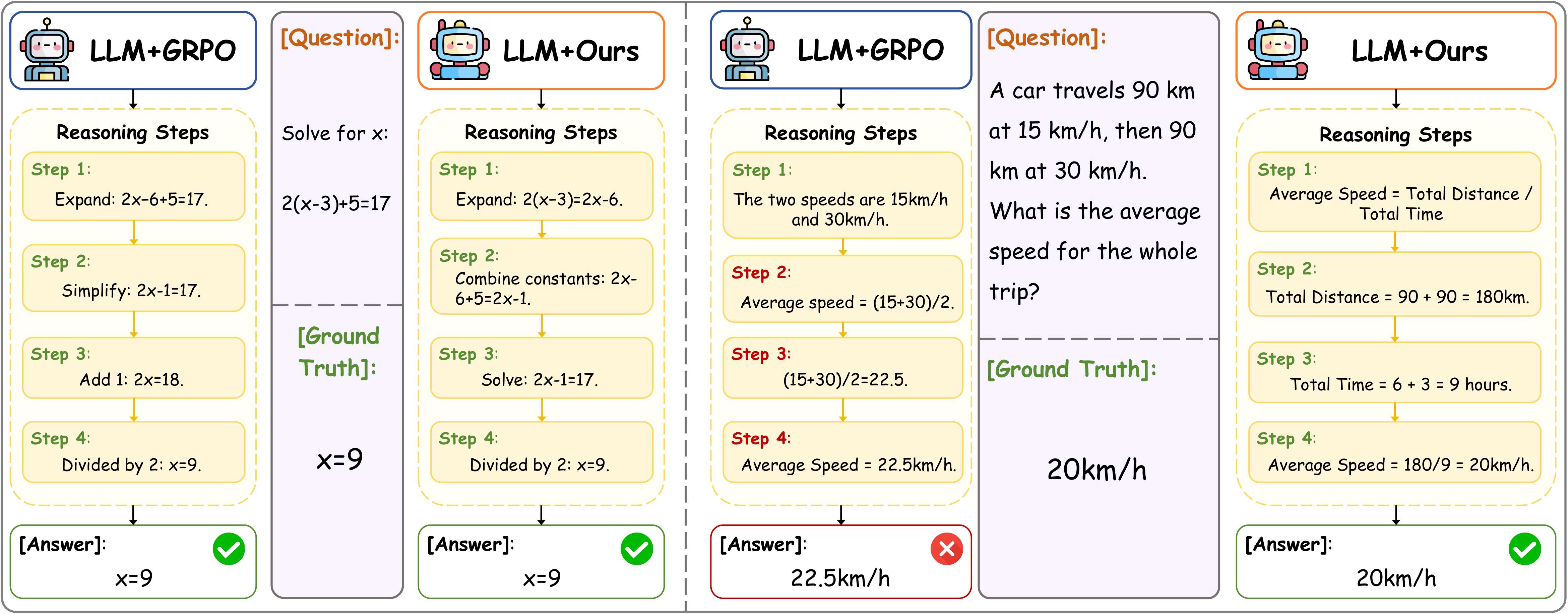}
        \caption{Qualitative Results.}
        \label{fig:vis}
    \end{minipage}
    \hfill
    \begin{minipage}[b]{0.49\linewidth}
        \centering
        \begin{subfigure}[b]{0.492\linewidth}
            \centering
            \includegraphics[width=\textwidth]{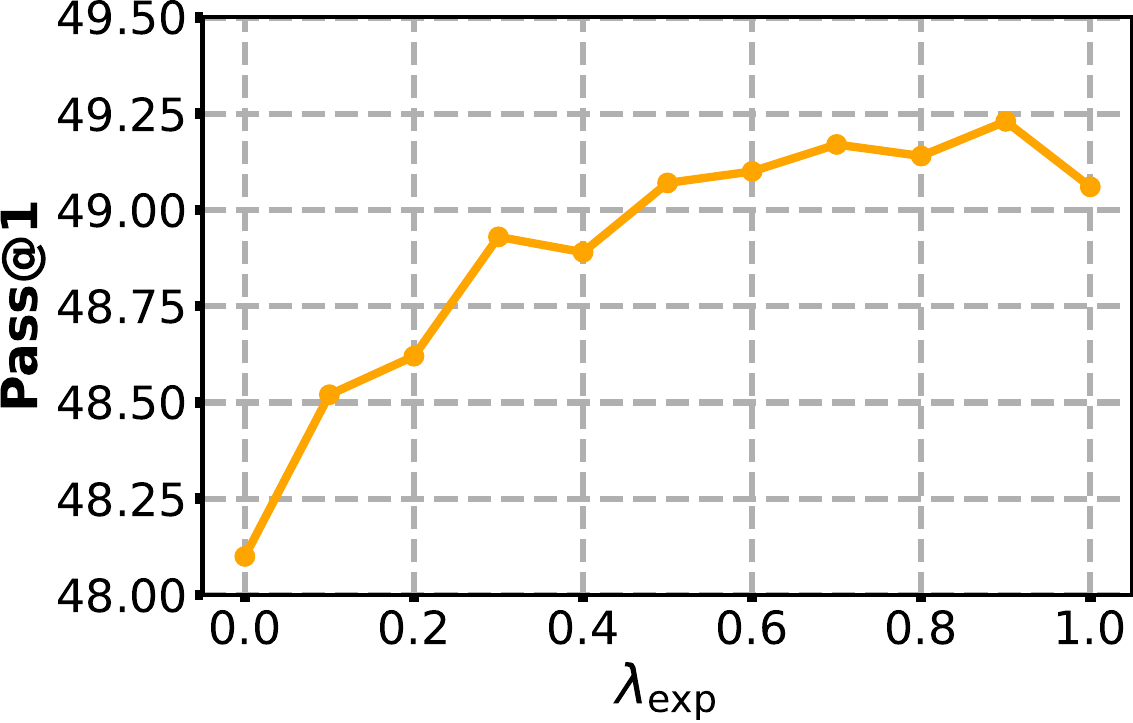}
            \caption{Evaluation of $\lambda_{\mathrm{exp}}$.}
            \vspace{-0.05in}
            \label{fig:norm_a}
        \end{subfigure}
        \hfill
        \begin{subfigure}[b]{0.488\linewidth}
            \centering
            \includegraphics[width=\textwidth]{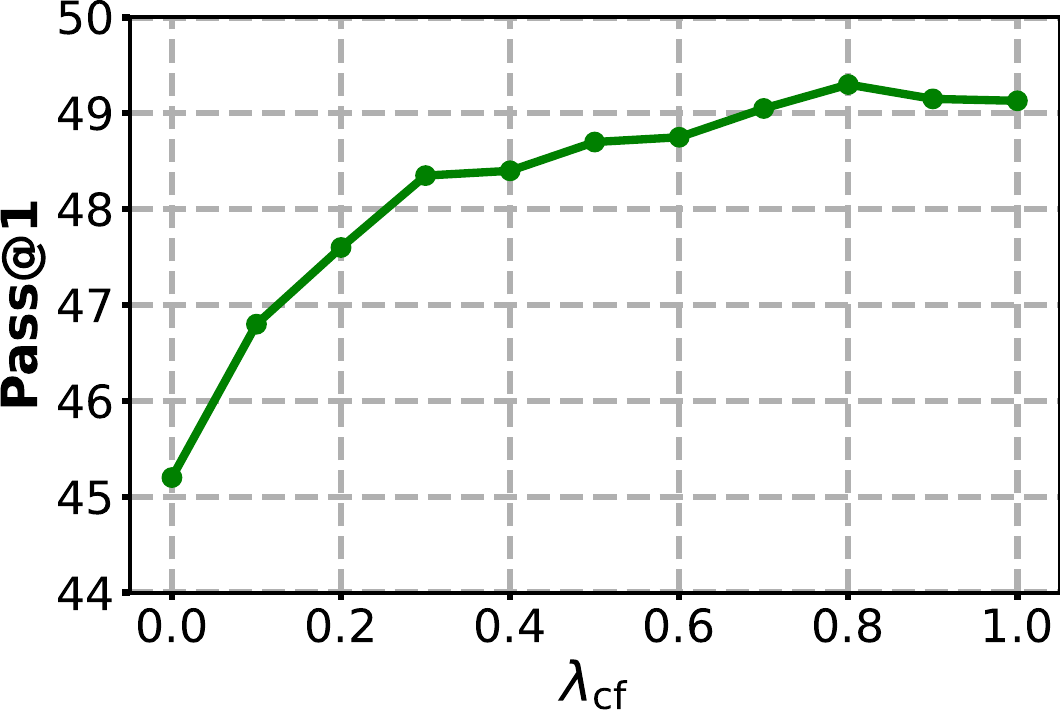}
            \caption{Evaluation of $\lambda_{\mathrm{cf}}$.}
            \vspace{-0.05in}
            \label{fig:norm_b}
        \end{subfigure}
        \caption{Ablation Studies.}
        \label{fig:ex_alba}
    \end{minipage}
    \vspace{-0.15in}
\end{figure}


\section{Experiments}
\label{sec:experiment}

\subsection{Experimental Settings}
\label{sec:ex_settings}
We evaluate on a range of reasoning benchmarks, including (i) math, i.e., AIME24-25, AMC, MATH500 \citep{hendrycks2021measuring}, MinervaMATH \citep{lewkowycz2022solving},  GSM8K \citep{cobbe2021training}, GSM-Plus \citep{li2024gsm}, (ii) coding, i.e., HumanEval \citep{chen2021evaluating}, (iii) commonsense reasoning, i.e., CommonsenseQA \citep{talmor2019commonsenseqa}, and (iv) formal logic reasoning, i.e., FOLIO \citep{han2024folio}; The base models cover different scales, e.g., 1.5B of DeepScaleR-1.5B-Preview and DeepSeek-R1-Distill-Qwen-1.5B, 7B of Qwen2.5-7B-Instruct and DeepSeek-R1-Distill-Qwen-7B, etc. We compare G$C^2$PO against (i) outcome-reward RL methods, e.g., GRPO \citep{shao2024deepseekmath}, Dr.GRPO \citep{liu2025understanding}, etc., and (ii) process-reward RL methods, e.g., GCPO \citep{gu2025group}, MRT \citep{qu2025optimizing}, etc.
Following \citep{qu2025optimizing,wang2025learning}, DeepScaleR-1.5B-Preview is fine-tuned on 919 AIME problems. DeepSeek-R1-Distill-Qwen-1.5B and 7B are fine-tuned on a random subset of 4,000 NuminaMath QA pairs \citep{li2024numinamath}. We use a learning rate of $1\times10^{-6}$, weight decay of 0.01, and batch size 256. 
The hyperparameters $\lambda_{\mathrm{exp}}$ and $\lambda_{\mathrm{cf}}$ are set to 0.9 and 0.8. All experiments are conducted on A100 GPU clusters.

\subsection{Results}
\label{sec:ex_results}

\textbf{Performance and Generalization Analyses.} 
We evaluate G$C^2$PO across all the benchmarks and base models, measuring both expected pass@1 and total token usage. 
As shown in \textbf{Table~\ref{tab:ex_1}}, \textbf{Figure \ref{fig_app:ex_token_usage}} and \textbf{Appendix E.2}, (i) G$C^2$PO achieves the best performance with fewer tokens, e.g., G$C^2$PO improves pass@1 by over 2.5\% against the baselines; (ii) G$C^2$PO consistently outperforms baselines on datasets whose distributions differ from the training data, demonstrating great generalization.

\textbf{Trade-off Performance.}
We (i) normalize the computational cost of GRPO to $1\times$ and measure total GPU hours across all the methods; and (ii) sample trajectories with a fixed token context window and truncate them at different budgets to evaluate performance.
\textbf{Figure~\ref{fig:efficiency_cost}} show that G$C^2$PO incurs only a modest overhead of $1.2\times$, while achieving the best accuracy with lower budget.

\textbf{Training Stability.}
Following common practice \citep{xiao2025bnpo,gu2025group}, we use the gradient norm as a proxy for policy variance. 
We also record the sampling ratio of our method compared to previous sequence-level methods. 
\textbf{Figure~\ref{fig:norm}} shows that G$C^2$PO exhibits more stability, with its gradient norm remaining nearly constant and the sampling ratio achieving no obvious spikes.

\textbf{Visualization Analysis.}
To illustrate how G$C^2$PO changes model behavior, we qualitatively compare models fine-tuned by different methods, visualizing their reasoning traces. \textbf{Figure \ref{fig:vis}} and \textbf{Appendix E.5} show that G$C^2$PO yields more structured reasoning: it decomposes problems into clear subgoals, revisits intermediate steps, and corrects inconsistencies; while GRPO often commits to a single rigid line of reasoning, even when the underlying reasoning is flawed.

\textbf{Ablation Study.} We conduct experiments on the hyperparameters $\lambda_{\mathrm{exp}}$ and $\lambda_{\mathrm{cf}}$. We search $\lambda_{\mathrm{exp}}$ and $\lambda_{\mathrm{cf}}$ over $[0, 1]$. Note that when they equal 0, we examine two ablated variants: (i) G$C^2$PO w/o $S_{\mathrm{exp}}$; (ii) G$C^2$PO w/o $S_{\rm nec}$; (iii) G$C^2$PO w/o $S_{\mathrm{sta}}\&S_{\mathrm{exp}}$ (also two terms of $R_{\mathrm{cf}}^{\star}$).
\textbf{Figure~\ref{fig:ex_alba}} (i) demonstrates the effectiveness of each component within G$C^2$PO. (ii) shows that the optimal result is at $\lambda_{\mathrm{exp}} = 0.9$ and $\lambda_{\mathrm{cf}} = 0.8$, which are our final configuration. See \textbf{Appendix E.4} for more results.

\textbf{More Results.} We also extend our experiments to domains, e.g., code generation, commonsense reasoning, formal logic reasoning, etc., and evaluate more aspects, e.g., the robustness of prompt segmentation and formatting compliance, the plug-and-play nature of G$C^2$PO, etc. The experimental results in \textbf{Appendix E} demonstrate the effectiveness of G$C^2$PO.


\section{Related Work}
Complex reasoning remains a challenging capability for LLMs \citep{jimenez2024swebenchlanguagemodelsresolve,snell2024scalingllmtesttimecompute}. To improve reasoning performance, several works \citep{snell2024scalingllmtesttimecompute,wang2025beyond,wang2025thoughtsplaceunderthinkingo1like,ye2025emergence} adopt RL-based post-training methods, which have become a key component underpinning frontier systems such as OpenAI o1 \citep{jaech2024openai} and DeepSeek R1 \citep{deepseekai2025deepseekr1}. Among these methods, GRPO \citep{shao2024deepseekmath} and its variants \citep{zhao2025geometric,gu2025group,liu2025understanding} have attracted particular interest, as they simplify optimization by using the group's average reward as an advantage baseline, making them well-suited for fine-tuning large-scale models. However, they rely on binary rewards that reflect only outcome correctness, ignoring the structure of the reasoning process \citep{wang2025learning,zheng2025group}. To address this issue, some works introduce process rewards to evaluate intermediate steps \citep{qu2025optimizing,zeng2025versaprm,zheng2025cold,zeng2025versaprm,wang2025learning,yuan2024free}. However, these methods still primarily target final correctness or rely on outcome labels for training, which may undermine the reasoning reliability and thus affect generalization. 
To fill this gap, in this paper, we analyze the reasoning of LLMs from a causal perspective. 
Different from other concurrent works that also leverage causality in LLMs \citep{zhang2025llm,vashishtha2025executable,yamin2025llms,lasheras2025interventional,chen2025counterbench,zheng2025cold}, which focus on specific biases (e.g., step length) or stress tests while overlooking the issue of reward misalignment, we focus on the ``lucky guess'' phenomenon and propose an episodic causal counterfactual reward for generalizable reasoning.
Our reward explicitly decouples reasoning quality from answer correctness, encouraging LLMs to acquire generalizable reasoning strategies. See \textbf{Appendix D} for more comparisons.


\section{Conclusion}
In this paper, we explore a critical bottleneck in RL-based post-training: the reward mechanisms of existing methods entangle process validity with final correctness, hindering the acquisition of generalizable reasoning. To overcome this, we establish a causal perspective that formalizes multi-candidate generation as a set of counterfactual experiments and propose Group Causal Counterfactual Policy Optimization (G$C^2$PO). It introduces a fine-grained reward mechanism to jointly evaluate the robustness and effectiveness of reasoning episodes under mechanism-preserving perturbations. This design decouples reasoning quality from outcome correctness: it provides positive supervision for logically sound reasoning steps, while suppressing spurious shortcuts and lucky guesses. By combining it with outcome reward, we build token-level advantage for optimization, encouraging LLMs to favor reasoning patterns that are both process-valid and answer-correct.
Extensive experiments across diverse benchmarks demonstrate the advantages of our method.

\section*{AI Usage Statement}

In this work, we used generative AI tools for grammar and language polishing. They were not involved in research ideation, methodology, analysis, or substantive writing. The authors take full responsibility for all contents of the paper.

\bibliography{reference}
\bibliographystyle{iclr2027_conference}

\end{document}

%% file: math_commands.tex
\usepackage{amsmath,amsfonts,bm}

\def\eqref#1{equation~\ref{#1}}

\def\1{\bm{1}}

\DeclareMathAlphabet{\mathsfit}{\encodingdefault}{\sfdefault}{m}{sl}
\SetMathAlphabet{\mathsfit}{bold}{\encodingdefault}{\sfdefault}{bx}{n}

